\documentclass[letterpaper]{article} 

\usepackage{amsmath}   
\usepackage{xcolor} 
\usepackage{multirow}  
\usepackage{booktabs}  
\usepackage{bibentry} 
\usepackage[caption=false]{subfig} 
\usepackage{amsfonts} 
\usepackage{makecell} 

\usepackage{aaai2026}  
\usepackage{times}  
\usepackage{helvet}  
\usepackage{courier}  
\usepackage[hyphens]{url}  
\usepackage{graphicx} 
\usepackage{natbib}  
\usepackage{caption} 
\usepackage{algorithm}
\usepackage{algorithmic}

\usepackage{booktabs}
\usepackage{newfloat}
\usepackage{listings}
\DeclareCaptionStyle{ruled}{labelfont=normalfont,labelsep=colon,strut=off} 
\floatstyle{ruled}
\newfloat{listing}{tb}{lst}{}
\floatname{listing}{Listing}
\title{Separate Aggregation of Split Network for Personalized Federated Learning}
\author{
    Yunseok Kang, Jaeyoung Song
}
\affiliations{
    \textsuperscript{}Department of Electronics Engineering, Pusan National University, Republic of Korea \\
    \{ys.kang, jsong\}@pusan.ac.kr
}

\begin{document}

\maketitle
\begin{abstract}
    Federated learning enables collaborative model training without sharing raw data, but its performance can degrade substantially under heterogeneous client data distributions. A single global model often cannot satisfy diverse client requirements, so personalized federated learning has therefore been explored to improve client specific performance while preserving global generalization. Existing PFL methods often face a fundamental tradeoff in which stronger global sharing can undermine local specialization, whereas stronger local adaptation can lead to overfitting under limited data, label imbalance, and missing class scenarios. In this work, we propose PGFedSplit, a personalized federated learning framework that improves both personalization and global generalization under severe client heterogeneity. PGFedSplit adopts a split architecture and performs adaptive aggregation scheduling tailored to the roles of different model components, enabling stable knowledge sharing while maintaining client specific adaptation. Each client further leverages a mixture of locally extracted representations and synthetic representations generated from server side Gaussian statistics, improving robustness under label imbalance and missing class conditions. Extensive experiments on Fashion MNIST, CIFAR 10, CIFAR 100, and Tiny ImageNet demonstrate consistent improvements over state of the art PFL methods, with stable convergence and superior personalization in highly heterogeneous settings.
\end{abstract}

\section{Introduction}
Federated Learning (FL) enables collaborative model training across distributed clients without sharing raw data \cite{mcmahan2017communication}. This paradigm is particularly attractive for on-device intelligence, where data are continuously generated but cannot be centralized due to privacy constraints. However, FL often degrades under statistical heterogeneity, because each client’s data distribution reflects individual usage patterns and environments \cite{li2020federated}. Such non-IID settings can produce biased global updates and unstable local optimization, especially when local data are scarce, label distributions are skewed, or certain classes are missing.

Personalized Federated Learning (PFL) aims to improve client-specific performance while still leveraging shared knowledge. A common design is to decompose a model into a shared representation layer and a personalized classification layer \cite{arivazhagan2019federated, collins2021exploiting}, motivated by the observation that earlier layers tend to learn transferable features, whereas later layers are more sensitive to client-specific decision boundaries \cite{bengio2013representation, xu2023personalized}. Despite its appeal, this decomposition induces a bias--variance trade-off in the personalized layer: client classifiers can become high-variance under scarce or skewed local data, yet overly aggressive sharing can bias them toward mismatched client distributions.

Many existing approaches reduce harmful cross-client influence through similarity-aware collaboration \cite{xu2023personalized, DBLP:journals/isci/ZhouZLHL25}, but personalization-layer learning can still remain high-variance and fragile under data scarcity, label imbalance, and missing-class conditions. We argue that a key reason is that different model components require different synchronization behaviors. Representation layers usually benefit from frequent global aggregation to learn stable and transferable features, whereas personalization layers can drift toward foreign distributions when synchronized too aggressively.

Motivated by this observation, we propose \textit{PGFedSplit}, a split PFL framework that jointly addresses variance and bias through layer-wise decoupled aggregation and a two-stage personalization design. Unlike prior split PFL methods that either avoid sharing the personalization layer or rely on a fixed synchronization schedule, PGFedSplit jointly determines \emph{when} to synchronize the personalization head and \emph{how} to adapt it locally using distribution-aware synthetic representations.

Specifically, PGFedSplit aggregates the representation layer in every communication round, while synchronizing the personalization layer through an adaptively scheduled periodic scheme. This periodic aggregation constructs a low-variance global personalization head by pooling evidence across clients, while avoiding overly frequent synchronization that can distort client-specific decision boundaries. After receiving the aggregated head, each client locally adapts it to its native distribution, which helps mitigate bias and preserve client specificity. In parallel, clients upload class-level representation statistics in every round, from which the server estimates global class-conditional Gaussian statistics. These statistics are then used to generate synthetic representations for personalization-head training according to each client’s local label distribution, improving robustness under label imbalance and missing-class conditions. Overall, PGFedSplit reduces classifier variance while controlling the bias induced by mismatched global sharing.

To further illustrate the sensitivity of the personalization layer to synchronization frequency, we study the interaction between the aggregation period and the local--global adaptation coefficient. Let the adapted personalization head be
\[
\phi_i^{\mathrm{adapt}} = \alpha \phi_i^{\mathrm{local}} + (1-\alpha)\phi^{\mathrm{global}},
\]
where $\alpha$ controls the contribution of the local and aggregated heads. For this analysis, we disable APA and replace the learned $\alpha_i^{(t)}$ with a fixed $\alpha$ shared across clients and synchronization rounds.

\begin{figure}[t!]
\centering
\includegraphics[width=0.72\linewidth]{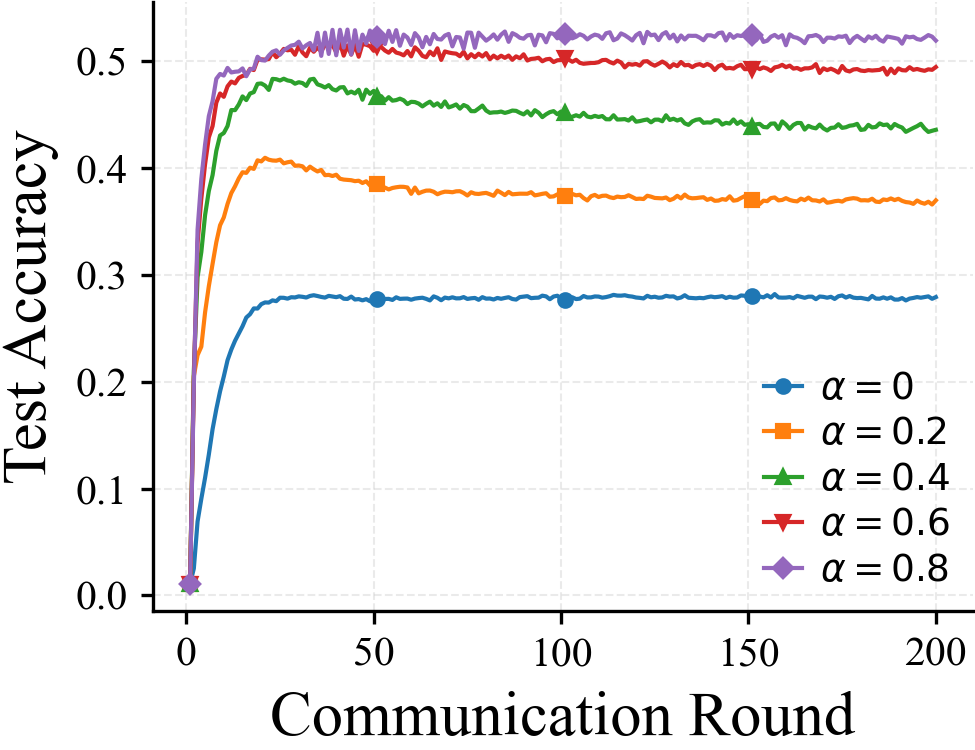}
\caption{Accuracy under different $\alpha$ when the personalization layer is aggregated every round.}
\label{fig:everyround_alpha}
\end{figure}

\begin{figure}[t!]
\centering
\includegraphics[width=0.48\linewidth]{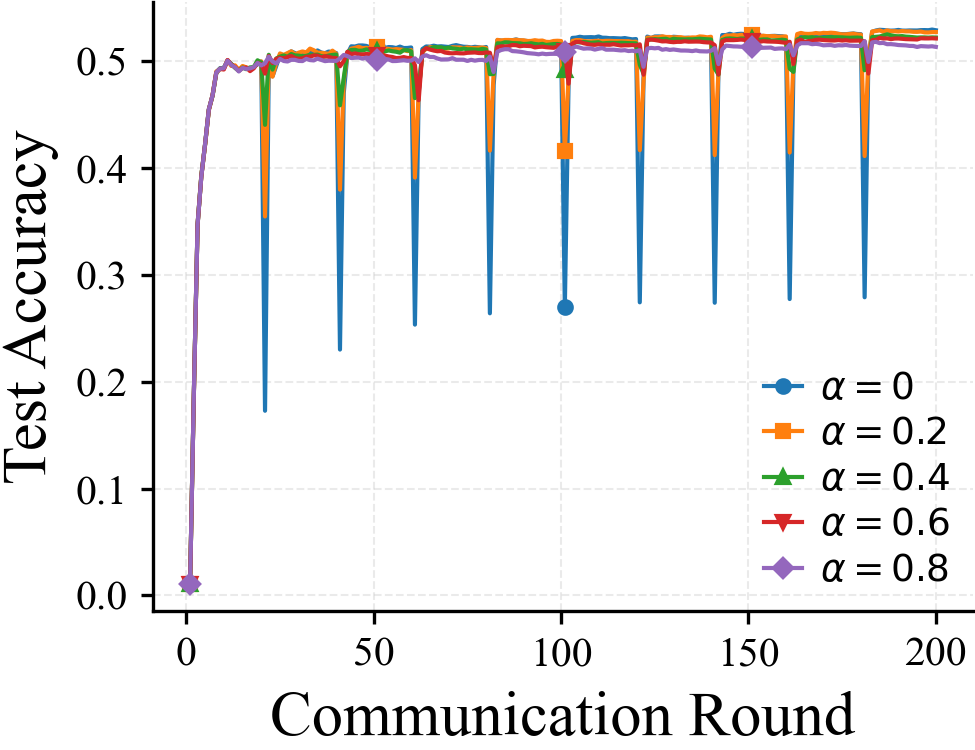} \hfill
\includegraphics[width=0.48\linewidth]{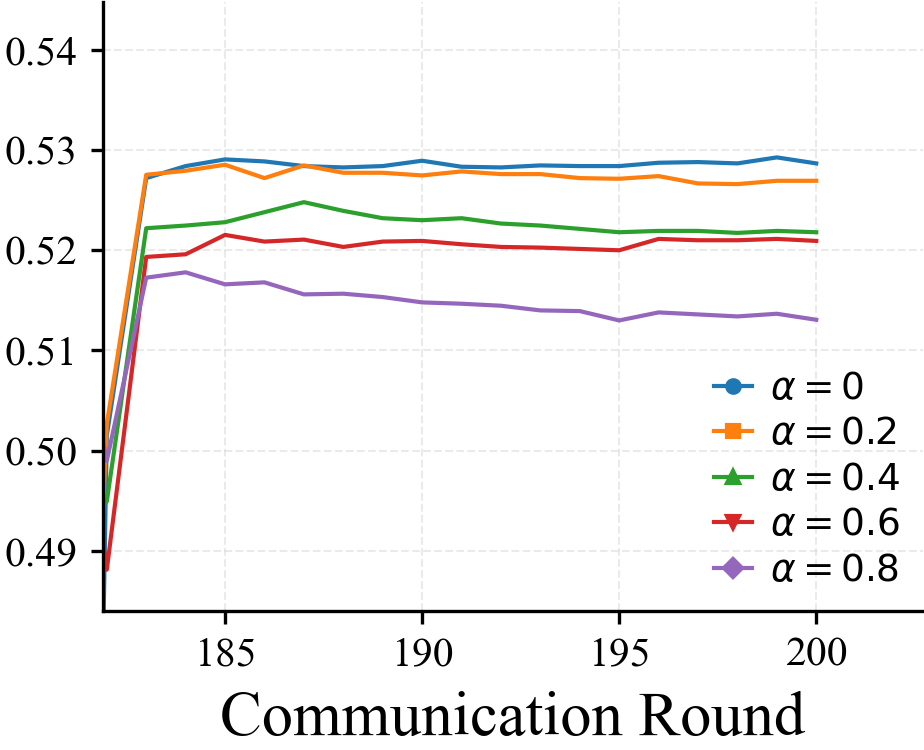}
\caption{Accuracy under different $\alpha$ when the personalization layer is aggregated every 20 rounds. Right: zoomed final convergence.}
\label{fig:round20_alpha}
\end{figure}

Figure~\ref{fig:everyround_alpha} shows the case where the personalization layer is aggregated at every communication round. In this regime, preserving the local classifier is critical: when $\alpha$ is too small, the aggregated head repeatedly overwrites local specialization and degrades performance. By contrast, Figure~\ref{fig:round20_alpha} shows the case where the personalization layer is aggregated every 20 rounds. With a longer interval, incorporating global information becomes more important, since the local classifier can drift toward narrow local objectives during extended local training. In this case, a moderate $\alpha$ improves final convergence, whereas overly strong global injection causes noticeable performance drops after aggregation.

These observations suggest that no single fixed synchronization rule is uniformly effective across regimes. Representation layers benefit from frequent communication because they should capture transferable features, whereas personalization layers should be synchronized more conservatively. Overly frequent synchronization harms local specificity, while overly aggressive correction after long local updates increases drift. This motivates our design of an adaptively scheduled personalization-head aggregation scheme combined with local re-adaptation.

Our main contributions are summarized as follows:
\begin{itemize}
    \item We propose a split PFL framework with layer-wise decoupled synchronization, aggregating the representation layer every round while adaptively scheduling the synchronization of the personalization layer.
    \item We introduce a two-stage personalization strategy that first constructs a low-variance global personalization head through adaptive periodic aggregation and then locally adapts it to mitigate bias and preserve client specificity.
    \item We further stabilize personalization-head training using distribution-aware synthetic representations generated from server-side class-conditional Gaussian statistics, and extensive experiments show consistent gains under severe heterogeneity, label imbalance, and missing-class settings.
\end{itemize}

\section{Related Works}

\subsection{Federated Learning}
Federated Learning (FL) enables collaborative model training across distributed clients without sharing raw data \cite{mcmahan2017communication}. A central challenge in FL is statistical heterogeneity across clients, which can lead to biased global updates and unstable convergence under non-IID data distributions \cite{li2020federated}. To improve robustness under such heterogeneity, prior work has explored optimization-based corrections such as proximal regularization \cite{li2020federated} as well as data-augmentation-based strategies \cite{yoon2021fedmix}.

Beyond optimization instability, practical FL systems may also suffer from representation-level heterogeneity, label imbalance, and missing-class scenarios, under which direct gradient or parameter averaging can become less effective. To alleviate this issue, prototype-based methods exchange class-level summaries instead of raw gradients or full model parameters. FedProto \cite{tan2022fedproto} aligns class-level knowledge through prototype exchange, while FedNH \cite{dai2023fednhAAAI} further improves robustness under class imbalance by exploiting the semantics and uniformity of prototypes. These methods demonstrate the value of class-level representation sharing under heterogeneity, but they do not explicitly address how personalized classifiers should be synchronized and adapted in split personalized federated learning.

\subsection{Personalized Federated Learning}
A single global model often cannot satisfy diverse client requirements, motivating Personalized Federated Learning (PFL) \cite{kairouz2021advances, fallah2020personalized, arivazhagan2019federated}. PFL aims to improve client-specific performance while still benefiting from collaborative knowledge sharing under heterogeneous client distributions. Existing approaches mainly differ in how they regulate collaboration, decompose model components, or modify the optimization objective.

\paragraph{Similarity-based personalization.}
One line of work reduces harmful cross-client influence by promoting collaboration among clients with similar data distributions. Clustering-based methods explicitly partition clients and learn cluster-specific models \cite{ghosh2020efficient, sattler2020clustered}. Beyond explicit clustering, several methods use client similarity to modulate aggregation or model combination. FedAMP \cite{huang2021personalized} performs similarity-aware aggregation, FedPAC \cite{xu2023personalized} adaptively weights personalization layers according to inter-client correlations, and FedFOMO \cite{zhang2021personalized} allows clients to selectively combine peer models based on local utility. These approaches mainly focus on \emph{which clients} should collaborate. By contrast, our work focuses on \emph{which model component} should be synchronized and \emph{how often}, with particular emphasis on the synchronization behavior of the personalization head.

\paragraph{Model decomposition.}
Another widely adopted direction separates transferable knowledge from client-specific decision rules by sharing only part of the model. FedPer \cite{arivazhagan2019federated} splits a network into shared representation layers and private personalization layers, sharing only the representation component. FedRep \cite{collins2021exploiting} further improves this paradigm by decoupling the training of different components and updating them separately. FedBABU \cite{oh2022fedbabu} also follows a partial-sharing strategy by first learning shared representations and then fine-tuning client-specific heads. More recently, FedGH \cite{yi2023fedgh} highlights that a shared prediction head can still be effective when it is properly designed, even when local feature extractors remain heterogeneous. These methods establish the importance of separating shared and private model components, but they typically either avoid synchronizing the personalization layer or treat it in a static manner. Our method instead adaptively schedules personalization-head synchronization and explicitly re-adapts the received head to each client’s local distribution.

\paragraph{Objective-based personalization.}
A complementary line of research improves personalization by modifying the learning objective or optimization process. MOCHA \cite{smith2017federated} formulates PFL as multi-task learning, while Per-FedAvg \cite{fallah2020personalized} adopts a meta-learning formulation. pFedMe \cite{t2020personalized} performs personalized optimization through Moreau envelopes, and APPLE \cite{luo2022adapt} adaptively balances global and local objectives. Ditto \cite{li2021ditto} learns local models while constraining them to remain close to a global model, APFL \cite{deng2020adaptive} combines global and local models via adaptive weighting, and FedALA \cite{zhang2023fedala} further extends this idea to finer-grained parameter-wise mixing. These methods improve personalization through objective design or adaptive mixing, but they do not explicitly exploit class-level representation statistics to stabilize personalization-head learning under label imbalance and missing-class conditions.

Overall, prior work has separately shown the benefits of similarity-aware collaboration, split model architectures, adaptive personalization objectives, and prototype-based knowledge sharing. Our work lies at the intersection of these directions: we study the synchronization bias--variance trade-off of the personalization head in split PFL, adapt its synchronization period online, and further stabilize local head learning using class-level Gaussian statistics derived from aggregated prototypes.

\section{Preliminary}
We first introduce the personalized federated learning setting considered in this work and then describe the split-model perspective that motivates our method.

\subsection{Problem Formulation for PFL}
We consider a personalized federated learning (PFL) system with one server and $K$ clients indexed by $i \in \mathcal{K}=\{1,2,\ldots,K\}$. Each client $i$ holds a local dataset $\mathcal{D}_i$, which is partitioned by label over the label set $\mathcal{L}$. We denote by $\mathcal{D}_{i,l}$ the subset of client $i$'s samples belonging to class $l \in \mathcal{L}$, so that
\[
\mathcal{D}_i = \bigcup_{l\in\mathcal{L}} \mathcal{D}_{i,l}.
\]
We further denote the dataset sizes by $D_i = |\mathcal{D}_i|$ and $D_{i,l}=|\mathcal{D}_{i,l}|$. Each local dataset consists of sample--label pairs,
\[
\mathcal{D}_i = \{(x_{i,k}, y_{i,k})\}_{k=1}^{D_i},
\]
where $x_{i,k}$ is the $k$th input and $y_{i,k}$ is its corresponding label.

Let $w_i$ denote the local model parameters of client $i$. The personalized federated objective is to optimize client-specific models while still benefiting from collaborative training across clients:
\[
\min_{w_1,\ldots,w_K} \frac{1}{K}\sum_{i=1}^{K} F_i(w_i;\mathcal{D}_i),
\]
where $F_i(w_i;\mathcal{D}_i)$ is the local loss of client $i$ computed from its own data. In each communication round, clients update their models using mini-batch stochastic gradient descent. For a mini-batch $\mathcal{B}_i \subseteq \mathcal{D}_i$, one local update step is written as
\begin{equation}
w_i' = w_i - \eta \nabla_w F_i(w_i;\mathcal{B}_i),
\end{equation}
where $\eta$ is the learning rate. This local update can be repeated for multiple epochs before server-side aggregation. In the following, we will further decompose each local model into shared representation and personalized components.

\subsection{Splitting a Deep Neural Network}
Deep neural networks consist of layers with different roles. Empirical studies suggest that earlier layers tend to learn general and transferable features, whereas deeper layers become increasingly task- or client-specific \cite{yosinski2014transferable, bau2017network, dorszewski2025colorsclassesemergenceconcepts}. This observation motivates split-model approaches such as FedPer \cite{arivazhagan2019federated} and FedRep \cite{collins2021exploiting}, which separate a model into a shared representation component and a client-specific personalization component.

Following this perspective, we interpret the earlier layers as \emph{representation layers}, which extract features that can be shared across clients, and the later layers as \emph{personalization layers}, which adapt predictions to the local data distribution. This decomposition is particularly natural in PFL, where transferable knowledge and client-specific decision rules should be handled differently.

Formally, let $f(\cdot;\theta_i^{(t)})$ denote the feature extractor induced by the representation layers of client $i$ at communication round $t$. For each class $l \in \mathcal{L}$, client $i$ summarizes its local class-wise representation by a prototype
\begin{equation}
\mathbf{p}_{i,l}^{(t)}
=
\frac{1}{D_{i,l}}
\sum_{x \in \mathcal{D}_{i,l}} f(x;\theta_i^{(t)}),
\end{equation}
which is the mean embedding of class-$l$ samples on client $i$.

Let $\mathcal{S}^{(t)} \subseteq \mathcal{K}$ denote the set of participating clients at round $t$, and define
\begin{equation}
\mathcal{S}_{l}^{(t)}
=
\{\, i \in \mathcal{S}^{(t)} \mid D_{i,l} > 0 \,\}.
\end{equation}
The server aggregates the uploaded local prototypes to obtain a global prototype for class $l$:
\begin{equation}
\bar{\mathbf{p}}_{l}^{(t)}
=
\frac{1}{|\mathcal{S}_{l}^{(t)}|}
\sum_{i \in \mathcal{S}_{l}^{(t)}} \mathbf{p}_{i,l}^{(t)}.
\end{equation}

These class-level representations provide a compact summary of distributed feature information and will later be used to guide both representation learning and personalization-head adaptation in our method.

\section{Proposed Method}

\begin{table}[t]
\centering
\caption{Main notation.}
\label{tab:notation}
\small
\setlength{\tabcolsep}{4pt}
\begin{tabular}{p{0.30\columnwidth} p{0.62\columnwidth}}
\toprule
Symbol & Meaning \\
\midrule
$w_i=\{\theta_i,\phi_i\}$ & local model of client $i$ \\
$\theta_i,\phi_i$ & representation and personalization layers \\
$\mathbf{p}_{i,l}^{(t)}$ & local prototype of class $l$ at client $i$ \\
$\bar{\mathbf{p}}_{l}^{(t)}$ & global prototype of class $l$ \\
$(\boldsymbol{\mu}_l^{(t)},\boldsymbol{\Sigma}_l^{(t)})$ & Gaussian statistics for class $l$ \\
$\bar{\phi}^{(t)}$ & aggregated personalization head delivered for local adaptation \\
$\bar{\phi}^{\mathrm{tmp}}$ & temporary aggregated personalization head stored at the server \\
$\alpha_i^{(t)}$ & mixing coefficient for the personalization head \\
$\bar{\alpha}^{\mathrm{prev}}$ & previous mean mixing coefficient used in APA \\
$\tau^{(t)}$ & aggregation interval for the personalization head \\
$s^{(t)}$ & elapsed rounds since the last personalization-head aggregation \\
$\Delta_i^{(t)}$ & adaptation gap of client $i$, i.e., elapsed rounds since it last adapted an aggregated personalization head \\
$\tilde{\mathcal{D}}_i^{(t)}$ & mixed training set for the personalization head \\
$\mathcal{S}^{(t)}$ & participating clients at round $t$ \\
$\mathcal{P}^{(t)}$ & classes for which global prototypes are available \\
$T_{\mathrm{KD}}$ & temperature for KL regularization \\
$R$ & total number of communication rounds \\
\bottomrule
\end{tabular}
\end{table}

\begin{figure*}[t]
    \centering
    \includegraphics[width=0.75\textwidth]{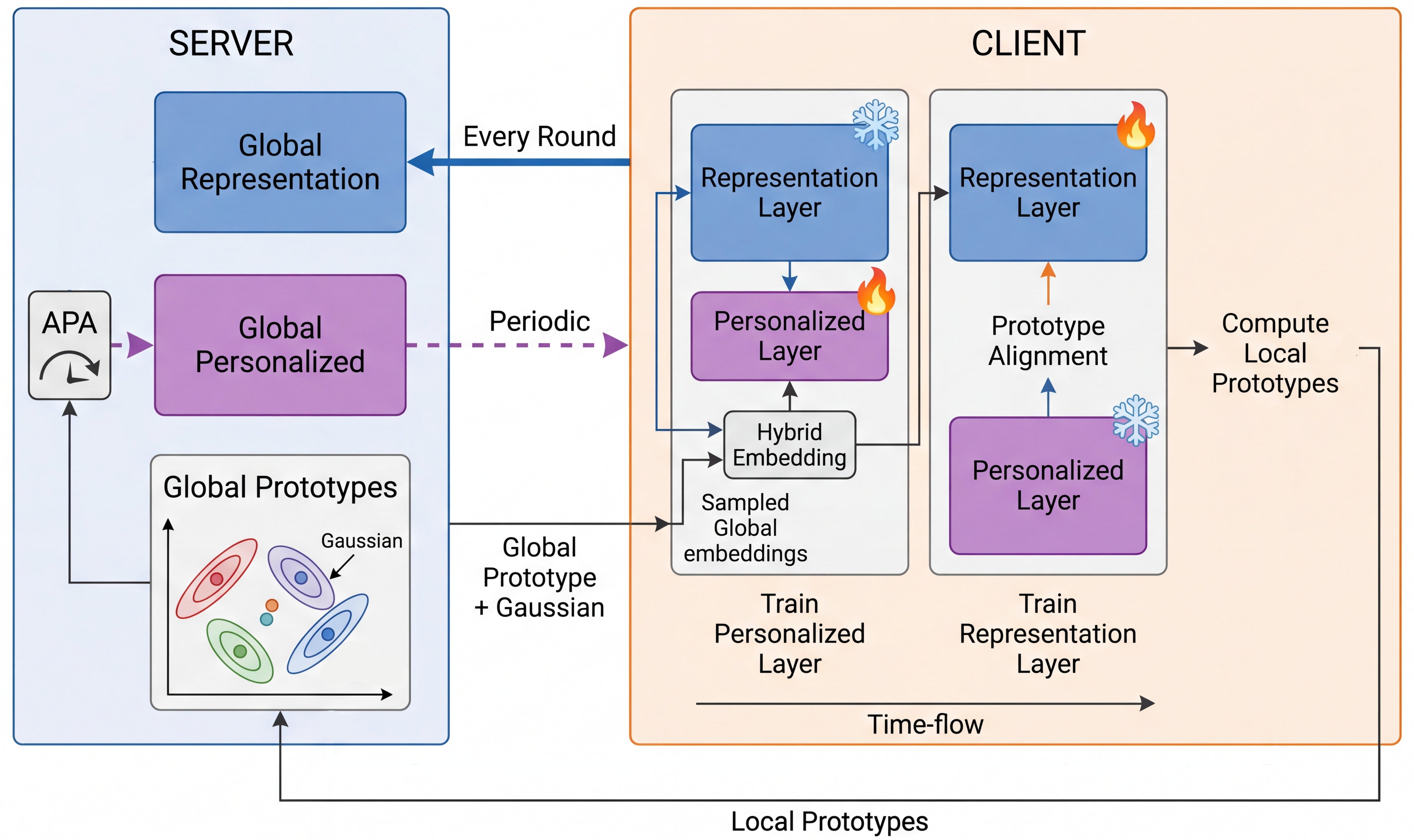}
    \caption{Overall PGFedSplit workflow: representation layers aggregate every round, personalization layers are synchronized periodically via APA, and global prototypes guide both personalization-head training (1) and representation regularization (2).}
    \label{fig:PGFedSplit_overview}
    \end{figure*}

    In this section, we present \textit{PGFedSplit}, which combines layer-wise decoupled synchronization with prototype-guided personalized learning. The proposed framework is designed to balance transferable global knowledge and client-specific adaptation under severe statistical heterogeneity. Figure~\ref{fig:PGFedSplit_overview} illustrates the overall workflow.
    
    \subsection{PGFedSplit Framework}
    
    In \textit{PGFedSplit}, each client improves its local model by handling the representation and personalization layers differently according to their roles. The learning process in each communication round consists of two coupled components: decoupled local training and periodic personalization-head aggregation.
    
    \paragraph{Decoupled local training.}
    We decompose the local model of client $i$ as
    \begin{equation}
    w_i := \{\theta_i,\phi_i\},
    \end{equation}
    where $\theta_i$ denotes the representation layers and $\phi_i$ denotes the personalization layers \cite{arivazhagan2019federated, collins2021exploiting}. Given a mini-batch $\mathcal{B}_i$, client $i$ first updates the personalization layers while freezing the representation layers:
    \begin{equation}
    \phi_i' = \phi_i - \eta_\phi \nabla_{\phi} F_i(\{\theta_i,\phi_i\}; \mathcal{B}_i),
    \end{equation}
    and then updates the representation layers while freezing the updated personalization layers:
    \begin{equation}
    \theta_i' = \theta_i - \eta_\theta \nabla_{\theta} F_i(\{\theta_i,\phi_i'\}; \mathcal{B}_i).
    \end{equation}
    This update order is motivated by the fact that the personalization layers are more directly affected by local label statistics and decision boundaries. Updating the personalization head first provides a sharper task-aligned supervisory signal for subsequent refinement of the shared representation.
    
    \subsubsection{Split Aggregation}
    Building on this decomposition, the server aggregates the representation layer in every communication round using data-size-weighted averaging, and broadcasts the updated representation to clients. Frequent synchronization is desirable for representation learning because it promotes stable and transferable features across heterogeneous clients.
    
    In contrast, the personalization layer is substantially more sensitive to client-specific bias. Direct and frequent averaging of personalization heads can therefore distort local decision boundaries and degrade client specificity. For this reason, prior work often either keeps the personalization layer fully local \cite{collins2021exploiting, arivazhagan2019federated} or applies similarity- or objective-aware mixing strategies \cite{zhang2023fedala, xu2023personalized, deng2020adaptive}. Our method instead synchronizes the personalization layer periodically and adapts the synchronization interval online.
    
    \subsubsection{Personalization Layer Adaptation}
    
    When a client receives an aggregated personalization head, it does not directly replace its local head. Instead, it combines the received head with its local personalization parameters to reduce bias while preserving locally specialized knowledge. Specifically, the personalization head is updated as
    \begin{equation}\label{eq:personalization_update}
    \phi_i^{(t)}
    =
    \alpha_i^{(t)}\phi_i^{(t-1)}
    +
    \bigl(1-\alpha_i^{(t)}\bigr)\bar{\phi}^{(t)},
    \end{equation}
    where $\bar{\phi}^{(t)}$ is the aggregated personalization head delivered for local adaptation and $\alpha_i^{(t)}\in[0,1]$ controls the relative contribution of local versus aggregated personalization.
    
    \paragraph{Adaptive Period Adjustment (APA).}
    To control the synchronization cadence of the personalization layer, the server performs adaptive period adjustment (APA) based on the mean mixing coefficient across clients. Let $\mathcal{C}^{(t)}$ denote the set of clients that update $\alpha$ at round $t$, and define
    \begin{equation}\label{eq:mean_alpha}
    \bar{\alpha}^{(t)}
    =
    \frac{1}{|\mathcal{C}^{(t)}|}
    \sum_{i\in\mathcal{C}^{(t)}} \alpha_i^{(t)}.
    \end{equation}
    Starting from an initial interval $\tau^{(0)}$, the server updates the aggregation interval by one step according to the change in the mean coefficient. If $\bar{\alpha}^{(t)}$ is larger than the previous mean $\bar{\alpha}^{\mathrm{prev}}$, the next interval is shortened by one round; if it is smaller, the interval is lengthened by one round; otherwise, the interval remains unchanged. In all cases, the updated interval is clipped to the range $[\tau_{\min},\tau_{\max}]$, i.e.,
    \begin{equation}\label{eq:tau_update}
    \begin{aligned}
    \tau^{(t+1)}
    =
    \operatorname{clip}_{[\tau_{\min},\tau_{\max}]}
    \Big(
    &\tau^{(t)}
    - \mathbf{1}\!\left[\bar{\alpha}^{(t)}>\bar{\alpha}^{\mathrm{prev}}\right]
    \\
    &+ \mathbf{1}\!\left[\bar{\alpha}^{(t)}<\bar{\alpha}^{\mathrm{prev}}\right]
    \Big).
    \end{aligned}
    \end{equation}
    Under this default schedule, a larger mean $\alpha$ leads to more frequent personalization-head synchronization, whereas a smaller mean $\alpha$ leads to less frequent synchronization.
    
    Let $s^{(t)}$ denote the number of rounds elapsed since the most recent personalization-head aggregation. The server increments this counter by one each round and triggers an aggregation when the counter reaches the current synchronization interval, after which it resets the counter to zero. The aggregated personalization head is stored temporarily and broadcast at the beginning of the next communication round before local training.
    
    \paragraph{Adaptation-gap-aware personalization-head adaptation.}
    A key challenge in personalized federated learning is that clients do not always receive and adapt an aggregated personalization head at the same frequency. If a client has not adapted an aggregated personalization head for many rounds, its local personalization head may become weakly aligned with the most recent global knowledge. To account for this effect, we introduce a client-specific adaptation gap
    \begin{equation}\label{eq:gap_def}
    \Delta_i^{(t)} = t - t_i^{\mathrm{last}},
    \end{equation}
    where $t_i^{\mathrm{last}}$ denotes the most recent round before $t$ in which client $i$ received and adapted an aggregated personalization head.
    
    We update $\alpha_i^{(t)}$ only when an aggregated personalization head is delivered. Let $z$ denote an embedding produced by the frozen representation layer, and let $g_{\phi}$ denote the classifier induced by personalization parameters. We define mixed logits for embedding $z$ as
    \begin{equation}\label{eq:mix_logits}
    \ell_{\alpha_i^{(t)}}(z)
    =
    \alpha_i^{(t)} g_{\phi_i}(z)
    +
    \bigl(1-\alpha_i^{(t)}\bigr) g_{\bar{\phi}}(z).
    \end{equation}
    Then, $\alpha_i^{(t)}$ is obtained by minimizing
    \begin{equation}\label{eq:alpha_obj}
    \begin{aligned}
    \alpha_i^{(t)} \in \arg\min_{\alpha\in[0,1]}
    \quad &
    \mathbb{E}_{(z,y)\sim \tilde{\mathcal{D}}_i^{(t)}}
    \bigl[
    \ell_{\mathrm{CE}}(\ell_\alpha(z),y)
    \bigr]
    \\
    &\quad
    + \beta \,\Delta_i^{(t)}\, \alpha^2
    D_{\mathrm{KL}}\!\bigl(p_i(z)\,\|\,p_g(z)\bigr),
    \end{aligned}
    \end{equation}
    where $\beta>0$ is a regularization coefficient. Here, $p_i(z)$ and $p_g(z)$ denote the temperature-scaled predictive distributions obtained by applying the softmax operator to $g_{\phi_i}(z)/T_{\mathrm{KD}}$ and $g_{\bar{\phi}}(z)/T_{\mathrm{KD}}$, respectively.
    
    The temperature-scaled KL term softens the predictive distributions of the local and aggregated personalization heads before measuring their discrepancy. A higher temperature reveals relative inter-class similarity structure beyond the top-1 class and yields a smoother alignment signal, which makes the optimization of $\alpha_i^{(t)}$ less sensitive to overly confident logits. This follows the standard rationale of knowledge distillation and is consistent with prior federated distillation-style knowledge transfer \cite{hinton2015distilling, li2019fedmd, lin2020ensemble}. Unless stated otherwise, we set the KL temperature to $T_{\mathrm{KD}}=1$.
    
    The adaptation-gap factor $\Delta_i^{(t)}$ makes this update explicitly client-aware. A larger $\Delta_i^{(t)}$ indicates that client $i$ has gone for a longer period without adapting an aggregated personalization head and is therefore more likely to benefit from stronger global correction at the next update. Consequently, clients with larger adaptation gaps are encouraged to place more weight on the aggregated personalization head, whereas clients that adapted more recently can retain stronger local specialization.
    
    \subsubsection{Prototype-Guided Learning}
    We now describe the prototype-guided component of our method in detail. After receiving the server-broadcast global prototypes $\{\bar{\mathbf{p}}_{l}^{(t)}\}$, each client uses them from two perspectives: representation refinement and personalization-head training.
    
    \paragraph{Representation refinement.}
    Clients update the representation layer $\theta_i$ while freezing the personalization layer $\phi_i$. Given a mini-batch $\mathcal{B}_i = \{(x_{i,k}, y_{i,k})\}_{k=1}^{|\mathcal{B}_i|}$ sampled from $\mathcal{D}_i$, client $i$ minimizes the following prototype-regularized loss:
    \begin{equation}\label{eq:prototype_loss}
    \begin{split}
    \mathcal{L}_{\mathrm{base}}(\theta_i^{(t)})
    &=
    \frac{1}{|\mathcal{B}_i|}
    \sum_{k=1}^{|\mathcal{B}_i|}
    \Big[
    \ell_{\mathrm{CE}}\big(\phi_i(\theta_i(x_{i,k})), y_{i,k}\big)
    \\
    &\qquad\qquad
    +
    \lambda \mathbf{1}_{\{y_{i,k}\in \mathcal{P}^{(t)}\}}
    \bigl\|
    \theta_i(x_{i,k})-\bar{\mathbf{p}}_{y_{i,k}}^{(t)}
    \bigr\|_2^2
    \Big],
    \end{split}
    \end{equation}
    where $\lambda$ controls the strength of prototype regularization, and $\mathcal{P}^{(t)}$ denotes the set of classes for which global prototypes are available. The indicator function ensures that the prototype regularization term is applied only when the corresponding global prototype is available. Consequently, the representation of a local sample is encouraged to stay close to the aggregated global prototype of its class, improving class-conditional alignment while avoiding unnecessary constraints on unsupported classes.
    
    \paragraph{Gaussian-Guided Personalized layer Tranining}
    To enable efficient training of the personalization head, each client constructs a mixed training set in the representation space by combining local feature embeddings with global synthetic embeddings. Since the number of classes is much smaller than the number of local samples, treating prototypes as deterministic copies can provide an overly concentrated supervisory signal. Instead, we generate global embeddings by sampling from class-wise Gaussian distributions estimated at the server.
    
    Along with the global prototype means $\{\bar{\mathbf{p}}_{l}^{(t)}\}$, the server broadcasts class-wise statistics $\{(\boldsymbol{\mu}_{l}^{(t)}, \boldsymbol{\Sigma}_{l}^{(t)})\}$, where $\boldsymbol{\mu}_{l}^{(t)}=\bar{\mathbf{p}}_{l}^{(t)}$ and $\boldsymbol{\Sigma}_{l}^{(t)}$ captures the dispersion of embeddings for class $l$ across clients. Each client forms a mixed training set for the personalization head by adding $N_{i,\mathrm{g}}^{(t)}$ sampled global embeddings to its $N_{i,\mathrm{l}}^{(t)}$ local embeddings, where the mixing ratio is controlled by a hyperparameter $r\in(0,1)$:
    \begin{equation}
    N_{i,\mathrm{g}}^{(t)} = \left\lceil \frac{r}{1-r} N_{i,\mathrm{l}}^{(t)} \right\rceil.
    \end{equation}
    The class label of each global sample is drawn with replacement according to the local label proportions $q_{i,l}=D_{i,l}/D_i$. Conditioned on a sampled class label $l$, the corresponding global embedding is generated as
    \begin{equation}\label{eq:gauss_sample}
    \tilde{\mathbf{p}}_{l,m}^{(t)} \sim
    \mathcal{N}\!\bigl(\boldsymbol{\mu}_{l}^{(t)},\ \gamma_{i,l}^{2}\boldsymbol{\Sigma}_{l}^{(t)}\bigr),
    \qquad
    m=1,\dots,N_{i,l}^{(t)},
    \end{equation}
    where $\gamma_{i,l}$ is a class-dependent variance scaling factor computed from the local label proportions, optionally normalized across classes.
    
    Each client embeds all local samples through the frozen representation layer to obtain local embeddings, and augments them with the sampled Gaussian embeddings. The resulting hybrid dataset is
    \begin{equation}\label{eq:aug_dataset}
    \tilde{\mathcal{D}}_i^{(t)}
    =
    \bigcup_{l \in \mathcal{L}}
    \left(
    \{(\theta_i(x),l)\mid(x,l)\in\mathcal{D}_{i,l}\}
    \cup
    \{(\tilde{\mathbf{p}}_{l,m}^{(t)},l)\}_{m=1}^{N_{i,l}^{(t)}}
    \right).
    \end{equation}
    After obtaining $\tilde{\mathcal{D}}_i^{(t)}$, client $i$ updates the personalization head $\phi_i$ using cross-entropy loss while keeping $\theta_i$ frozen. Consequently, training requires only forward and backward passes through $\phi_i$, yielding low computational overhead while incorporating both local characteristics and global knowledge.
    
    Algorithms~\ref{alg:server} and \ref{alg:client} summarize the server-side and client-side procedures of PGFedSplit.
    
    \begin{algorithm}[t]
    \caption{PGFedSplit: Server-Side Procedure}
    \label{alg:server}
    \begin{algorithmic}[1]
    \STATE \textbf{Initialize:} $\bar{\theta}^{(0)}$, $\{\bar{\mathbf{p}}_l^{(0)}\}_{l\in\mathcal{L}}$,
    $\{(\boldsymbol{\mu}_l^{(0)}, \boldsymbol{\Sigma}_l^{(0)})\}_{l\in\mathcal{L}}$,
    $\tau^{(0)}$, $s \leftarrow 0$, $\bar{\alpha}^{\mathrm{prev}} \leftarrow 0$
    \STATE \hspace{1.2cm} temporary aggregated personalization head $\bar{\phi}^{\mathrm{tmp}} \leftarrow \emptyset$
    
    \FOR{$t = 0,1,\dots,R-1$}
        \STATE Select participating clients $\mathcal{S}^{(t)} \subseteq \mathcal{K}$
        \STATE Broadcast $\bar{\theta}^{(t)}$, $\{\bar{\mathbf{p}}_l^{(t)}\}_{l\in\mathcal{L}}$, and $\{(\boldsymbol{\mu}_l^{(t)}, \boldsymbol{\Sigma}_l^{(t)})\}_{l\in\mathcal{L}}$ to all clients
        \IF{$\bar{\phi}^{\mathrm{tmp}} \neq \emptyset$}
            \STATE Broadcast $\bar{\phi}^{\mathrm{tmp}}$ to all clients
        \ENDIF
    
        \STATE Receive $(\theta_i^{(t+1)}, \phi_i^{(t+1)}, \{\mathbf{p}_{i,l}^{(t+1)}\}_l)$ from each $i \in \mathcal{S}^{(t)}$
    
        \STATE Aggregate representation layers to obtain $\bar{\theta}^{(t+1)}$
        \STATE Aggregate local prototypes to obtain $\{\bar{\mathbf{p}}_l^{(t+1)}\}_{l\in\mathcal{L}}$
        \STATE Estimate Gaussian statistics $\{(\boldsymbol{\mu}_l^{(t+1)}, \boldsymbol{\Sigma}_l^{(t+1)})\}_{l\in\mathcal{L}}$
    
        \IF{$\bar{\phi}^{\mathrm{tmp}} \neq \emptyset$}
            \STATE Receive $\alpha_i^{(t)}$ from clients in $\mathcal{S}^{(t)}$ that updated $\alpha$
            \STATE Compute $\bar{\alpha}^{(t)}$ using Eq.~\eqref{eq:mean_alpha}
            \STATE Update $\tau^{(t+1)}$ using Eq.~\eqref{eq:tau_update}
            \STATE $\bar{\alpha}^{\mathrm{prev}} \leftarrow \bar{\alpha}^{(t)}$
            \STATE $\bar{\phi}^{\mathrm{tmp}} \leftarrow \emptyset$
        \ELSE
            \STATE $\tau^{(t+1)} \leftarrow \tau^{(t)}$
        \ENDIF
    
        \STATE $s \leftarrow s + 1$
        \IF{$s \ge \tau^{(t+1)}$}
            \STATE Aggregate personalization heads from $\mathcal{S}^{(t)}$ to obtain $\bar{\phi}^{(t+1)}$
            \STATE Store $\bar{\phi}^{(t+1)}$ in $\bar{\phi}^{\mathrm{tmp}}$
            \STATE $s \leftarrow 0$
        \ENDIF
    \ENDFOR
    \end{algorithmic}
    \end{algorithm}
    
    \begin{algorithm}[t]
    \caption{PGFedSplit: Client-Side Local Update}
    \label{alg:client}
    \begin{algorithmic}[1]
    \REQUIRE $\bar{\theta}^{(t)}$, $\{\bar{\mathbf{p}}_l^{(t)}\}_{l\in\mathcal{L}}$, $\{(\boldsymbol{\mu}_l^{(t)}, \boldsymbol{\Sigma}_l^{(t)})\}_{l\in\mathcal{L}}$, optional $\bar{\phi}^{(t)}$
    \STATE Set $\theta_i^{(t)} \leftarrow \bar{\theta}^{(t)}$
    
    \IF{$\bar{\phi}^{(t)}$ is received}
        \STATE Compute adaptation gap $\Delta_i^{(t)}$ using Eq.~\eqref{eq:gap_def}
        \STATE Compute $\alpha_i^{(t)}$ using Eq.~\eqref{eq:alpha_obj}
        \STATE Update the personalization head $\phi_i^{(t)}$ using Eq.~\eqref{eq:personalization_update}
    \ENDIF
    
    \STATE Construct $\tilde{\mathcal{D}}_i^{(t)}$ using Eq.~\eqref{eq:gauss_sample} and Eq.~\eqref{eq:aug_dataset}
    \STATE Update the personalization head $\phi_i$ on $\tilde{\mathcal{D}}_i^{(t)}$ while freezing $\theta_i$
    \STATE Update the representation layers $\theta_i$ using Eq.~\eqref{eq:prototype_loss} while freezing $\phi_i$
    \STATE Compute local prototypes $\{\mathbf{p}_{i,l}^{(t+1)}\}_{l\in\mathcal{L}}$
    \STATE Upload $(\theta_i^{(t+1)}, \phi_i^{(t+1)}, \{\mathbf{p}_{i,l}^{(t+1)}\}_l)$ to the server
    \IF{$\bar{\phi}^{(t)}$ is received}
        \STATE Upload $\alpha_i^{(t)}$ to the server
    \ENDIF
    \end{algorithmic}
    \end{algorithm}

	\section{Experiments}

	We conduct experiments for image classification using the Fashion‑MNIST (F‑MNIST) \cite{xiao2017}, CIFAR10/100 \cite{krizhevsky2009learning}, and Tiny-ImageNet (T-ImageNet) \cite{le2015tiny} datasets. For all image classification tasks except T‑ImageNet, we employ a 4‑layer CNN \cite{mcmahan2017communication}, whereas for T-ImageNet, we use a ResNet-18 model \cite{he2016residual}.
    We treat the final classifier head as the personalized layer and regard the remaining network as the representation layer.
	In our experiments, we consider a scenario where local dataset is distributed according to the Dirichlet distribution \cite{yurochkin2019,he2020group} with parameter $\beta$.
	As \(\beta\) decreases, the distribution becomes more skewed, thereby increasing the degree of heterogeneity among client datasets. In our experiments, $\beta \in \{0.1, 1\}$ is used to model high and low heterogeneous environment, respectively.
    In addition to that, we conduct experiments on pathological distribution \cite{mcmahan2017communication,shamsian2021} and results are represented in Technical Appendix.
	
    
    We set the number of clients to $K=20$, with all clients participating in every communication round unless otherwise specified. We use a learning rate of $\eta = 0.005$ and batch size $B = 64$. Each client partitions its dataset into 75\% training data and 25\% test data to evaluate local model performance. In all experiments, we set $\lambda = 5$.
    
    For the personalization layer, the aggregation interval is initialized as $\tau^{(0)}=5$ and updated online by APA unless otherwise specified. The local--global blending coefficient $\alpha_i^{(t)}$ is also learned online during training; hence, the main results do not rely on a single fixed $\alpha$. For the mixed representation set used to train the personalization head, we set the global-sample ratio to $r=0.5$ unless otherwise specified.
    
    We train all datasets for a total of 200 communication rounds, with five local epochs in each round. Throughout training, each local model is evaluated on its corresponding client's test set. We report the mean top-1 accuracy across clients over three independent runs with different random seeds, following the evaluation protocol of \cite{zhang2023fedala}.
    	

    \subsection{Performance Comparison}

    \begin{table*}[t]
      \centering
      \caption{Comparison of Top-1 test accuracy (\%) in the Dirichlet heterogeneous setting.}
      \label{tab:dirichlet_heterogeneous}
      \begin{normalsize}
      \begin{tabular}{lcccccccccc}
        \toprule
        \multirow{2}{*}{\textbf{Method}} &
        \multicolumn{2}{c}{\textbf{F-MNIST}} &
        \multicolumn{2}{c}{\textbf{CIFAR10}} &
        \multicolumn{2}{c}{\textbf{CIFAR100}} &
        \multicolumn{2}{c}{\textbf{T-ImageNet}} &
        \multicolumn{2}{c}{\textbf{Average}} \\
        & $\beta=0.1$ & $\beta=1$
        & $\beta=0.1$ & $\beta=1$
        & $\beta=0.1$ & $\beta=1$
        & $\beta=0.1$ & $\beta=1$
        & $\beta=0.1$ & $\beta=1$ \\
        \midrule
        Local        & 97.18 & 88.90 & 88.77 & 60.70 & 46.72 & 19.25 & 30.98 & 23.61 & 65.91 & 48.12 \\
        \midrule
        FedAvg       & 83.24 & 90.26 & 45.77 & 64.73 & 27.38 & 30.56 & 10.19 & 11.06 & 41.64 & 49.15 \\
        FedProx      & 83.24 & 90.27 & 45.75 & 64.69 & 27.43 & 30.61 & 9.90  & 10.34 & 41.58 & 48.98 \\
        \midrule
        FedAvg+FT    & 97.49 & 92.15 & 89.91 & 72.83 & 54.42 & 33.86 & 36.48 & 29.03 & 69.58 & 56.97 \\
        FedProx+FT   & 97.50 & 92.16 & 89.90 & 72.90 & 54.33 & 33.87 & 36.26 & 28.26 & 69.50 & 56.80 \\
        \midrule
        FedAMP       & 97.17 & 88.94 & 88.76 & 60.64 & 46.84 & 19.10 & 31.15 & 23.65 & 65.98 & 48.08 \\
        FedFomo      & 97.01 & 90.11 & 88.18 & 62.23 & 44.41 & 17.88 & 28.97 & 24.25 & 64.64 & 48.62 \\
        FedBABU      & 97.21 & 89.53 & 89.00 & 69.82 & 50.89 & 30.57 & \textbf{38.62} & 30.28 & 68.93 & 55.05 \\
        FedPer       & 97.44 & 90.80 & 89.13 & 66.66 & 46.28 & 21.17 & 34.39 & 24.71 & 66.81 & 50.84 \\
        FedRep       & 97.49 & 91.02 & 89.48 & 67.80 & 48.56 & 20.51 & 36.97 & 28.15 & 68.12 & 51.87 \\
        FedALA       & 97.51 & 92.75 & 89.94 & 73.55 & 54.41 & 34.76 & 34.93 & 19.38 & 69.20 & 55.11 \\
        FedPAC       & 97.41 & 92.93 & 88.37 & \textbf{77.25} & 63.13 & 41.82 & 35.20 & 26.29 & 71.03 & 59.57 \\
        FedGH        & 96.04 & 88.95 & 87.19 & 60.37 & 45.99 & 19.17 & 30.04 & 21.04 & 64.81 & 47.38 \\
        \textbf{PGFedSplit} & \textbf{97.62} & \textbf{92.84} & \textbf{90.57} & 77.14 & \textbf{64.41} & \textbf{43.17} & 38.45 & \textbf{30.91} & \textbf{72.76} & \textbf{61.02} \\
        \bottomrule
      \end{tabular}
      \end{normalsize}
    \end{table*}

	We compare the performance of PGFedSplit with existing strategy, including \textit{Local}, where each client trains its model locally without any exchange of updates. As comparing methods, traditional FL approaches that learn a single global model, including FedAvg \cite{mcmahan2017communication}, FedProx \cite{li2020federated} are considered. Additionally, we compare PGFedSplit with FedAvg-FT and FedProx-FT, which perform fine-tuning the global with local data model after standard FL training. For fine-tuning schemes, each client updates global model with $25$ epochs locally after all rounds. Moreover, PFL counterparts including FedPer \cite{arivazhagan2019federated}, FedRep \cite{collins2021exploiting}, FedPAC \cite{xu2023personalized}, FedAMP \cite{huang2021personalized}, FedFomo \cite{zhang2021personalized}, FedBABU \cite{oh2022fedbabu}, FedGH \cite{yi2023fedgh}, and FedALA \cite{zhang2023fedala} are compared.

    
	As shown in Table~\ref{tab:dirichlet_heterogeneous} In highly heterogeneous environment which corresponds to $\beta = 0.1$ single global model approaches exhibit relatively poor performance compared to other algorithms. However, for $\beta = 1$, where data distributions across clients are relatively uniform, single global model approaches perform comparatively well by benefiting from global aggregation. Nevertheless, these methods still underperform compared to fine-tuning and other PFL algorithms. Fine-tuning effectively balances a generalized global model with local adaptation to handle varying data distributions.
    However, among PFL algorithms, FedAMP's similarity-based method shows relatively lower performance, especially when $\beta=1$, as the advantage of similarity diminishes.  
    FedBABU, which fixes the biased heads during training and fine-tunes them after training, brings advantages in model generalization and shows overall good performance. FedPer and FedRep trained personalized layers independently while exchanging representation layers, effectively share meaningful information even when label distributions were highly divergent. FedALA also achieves strong performance across various datasets by adaptively applying local aggregation, although necessitating adjustment of aggregation constant over rounds. FedPAC applies label-based embeddings exclusively in the representation layer, excelling on CIFAR-10 but underperforming on T-ImageNet underscoring the need to carefully handle diverse label spaces.
    Notably, PGFedSplit consistently demonstrates the highest or near-highest accuracy across diverse and heterogeneous data distributions, proving its robustness.

    \subsection{Scalability}
    \begin{table}[t]
      \centering
      \begin{normalsize}
      \begin{tabular}{lccc}
        \toprule
        \textbf{Method} & \textbf{CIFAR10} & \textbf{CIFAR100} & \textbf{T-ImageNet} \\
        \midrule
        Local        & 84.15 & 40.30 & 28.14 \\
        \midrule
        FedAvg       & 43.83 & 17.88 & 9.70 \\
        FedProx      & 44.36 & 17.96 & 9.69 \\
        \midrule
        FedAvg+FT    & 79.57 & 40.82 & 32.49 \\
        FedProx+FT   & 79.68 & 40.76 & 32.54 \\
        \midrule
        FedAMP       & 84.42 & 40.48 & 28.26 \\
        FedFomo      & 83.53 & 37.82 & 27.30 \\
        FedBABU      & 83.31 & 40.46 & 35.81 \\
        FedPer       & 85.30 & 40.18 & 33.55 \\
        FedRep       & 85.09 & 39.69 & 34.57 \\
        FedALA       & 85.30 & 43.83 & 24.84 \\
        FedPAC       & 83.11 & 41.23 & 34.21 \\
        FedGH        & 77.84 & 35.35 & 27.20 \\
        \midrule
        \textbf{PGFedSplit} & \textbf{86.75} & \textbf{45.13} & \textbf{36.64} \\
        \bottomrule
      \end{tabular}
      \caption{Comparison of Top-1 test accuracy (\%) for $K=100$, participation rate $=0.3$, Dir($\beta=0.1$)).}
        \label{tab:scalability_dirichlet0.1}
      \end{normalsize}
    \end{table}
    
	Table~\ref{tab:scalability_dirichlet0.1} presents experimental results when we have $K=100$ clients and 30\% of the clients participate in each round. This partial participation scenario more accurately reflects real world FL scenario, compared to the ideal case in which all clients contribute in every round. 
    
     As shown in Table~\ref{tab:scalability_dirichlet0.1}, PGFedSplit consistently achieves top or near top accuracy across different datasets and partition types. When large number of clients collaborate, depending on the participation ratio, PGFedSplit can achieve high performance by adjusting aggregation period of personalization layer. For $K=100$ case, we set $\tau = 20$ to enable more local training on the personalization layer. 

    In particular, when small number of clients are participating in every round, longer period of aggregation for personalization layer is preferred. As aggregation period becomes longer, majority of clients can contribute in between consecutive aggregations, which leads to a well-updated personalization layer.

    \subsection{Model Generalization}
    \begin{figure}[h!]
      \centering
      \includegraphics[scale=0.499]{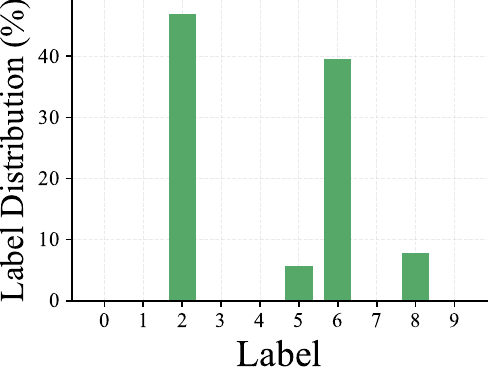}
      \hfill
      \includegraphics[scale=0.446]{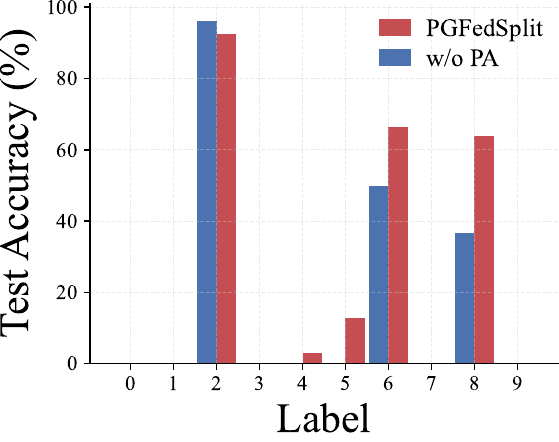}
      
      \caption{
    Effect of personalized layer aggregation on model generalization (CIFAR10, Dir($\beta=0.1$)).
    Left: Client data distribution. Right: Label-wise test accuracy. 
    }
      \label{fig:headagg}
    \end{figure}
    To demonstrate that aggregating personalized layers enhances model generalization, Figure~\ref{fig:headagg} presents test accuracy per label evaluated on a uniformly distributed test dataset for a chosen client.
    The results reveal that aggregating personalized layers slightly decreases the accuracy for the label representing the majority of data held by the client. However, it substantially improves accuracy for labels with fewer training samples. Moreover, we observed a slight but meaningful improvement in accuracy for label 4, which was not originally present in the client's dataset. This indicates enhanced generalization.
    In contrast, without aggregating personalized layers, the model tends to overfit the dominant labels, achieving high accuracy on those labels but significantly lower accuracy on underrepresented labels. Although this may have minimal impact when testing data distributions closely match the client's training data, performance degradation could become pronounced when facing different or more diverse test data distributions. These observations demonstrate that our proposed aggregation of personalized layers effectively enhances the generalization capacity of the model.

    \subsection{Ablation Study}
    
    We conduct an ablation study 
    to highlight the contributions of the two main components in our framework, we compare the following variants: \textbf{PGFedSplit (proposed)}, \textbf{w/o APA}, which removes adaptive period adjustment and uses a fixed personalization-head aggregation interval; \textbf{w/o GAU}, which removes Gaussian-guided head training and trains the personalization head without Gaussian-sampled synthetic feature representations; and \textbf{w/o \{APA, GAU\}}, which removes both components simultaneously.
    
    \begin{figure}[t]
    \centering
    \includegraphics[width=0.92\linewidth]{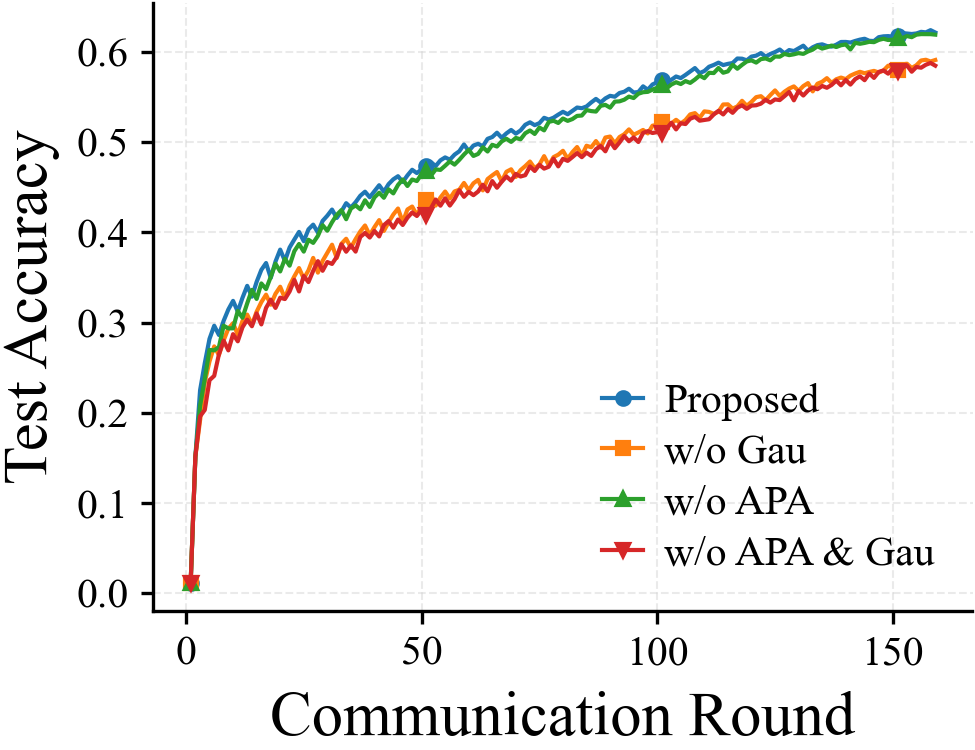}
    \caption{Ablation study on CIFAR-100 under Dirichlet $\beta=0.1$.}
    \label{fig:ablation_main}
    \end{figure}
    
    As shown in Figure~\ref{fig:ablation_main}, the proposed PGFedSplit consistently achieves the best performance throughout training, demonstrating the effectiveness of combining adaptive personalization-head synchronization with Gaussian-guided feature-space augmentation. Among the ablated variants, removing GAU leads to the most noticeable degradation in convergence speed and final accuracy, indicating that Gaussian-sampled synthetic feature representations provide a particularly strong supervisory signal under severe heterogeneity, label imbalance, and missing-class conditions.
    
    We also observe that removing APA causes a clear performance drop compared with the full model, confirming that adaptively adjusting the synchronization interval of the personalization head is beneficial over a fixed scheduling strategy. Although its effect is less pronounced than that of GAU in this experiment, APA still provides a meaningful improvement by better balancing local specialization and global correction during training.
    
    Finally, the \textbf{w/o \{APA, GAU\}} variant performs the worst among all compared settings. This result suggests that APA and GAU are complementary rather than redundant: GAU improves the efficiency and stability of personalization-head learning, while APA improves how and when global personalization information is incorporated. Taken together, these results support the design of PGFedSplit and show that each proposed component contributes meaningfully to the final performance.

    \section{Conclusion}
    In this paper, we propose PGFedSplit, a PFL strategy that partitions a neural network into representation layers and personalized layers, each aggregated at different frequencies.
    The shared representation layers are aggregated every round to enhance capturing meaningful features, while the personalized layers are intentionally aggregated less frequently to improve personalization.
    In particular, the personalized layers are trained on combined prototype and local data to further improve local adaptation.
    Our experimental results demonstrate that PGFedSplit significantly improves performance and generalization in heterogeneous environments. Moreover, PGFedsplit shows robustness in scalability. We believe PGFedSplit provides a promising approach toward efficient and robust personalized federated learning in real-world applications.

\bibliography{references}
\end{document}